\documentclass[wcp]{jmlr}

\usepackage{booktabs}

\makeatletter
\let \@jmlrpages \@empty

\title[ A Layer-wise Study]{ A Layer-wise Study of How Style Rewriting Breaks the Calibration of Zero-Shot LLM Fake-News Detectors}

\author{\Name{Prabhjot Singh} \Email{prabhjot\_s@cs.iitr.ac.in}\\
\addr Department of Computer Science, IIT Roorkee
\AND
\Name{Praney Goyal} \Email{praneyygoyal@gmail.com}\\
\addr Independent Researcher
\AND
\Name{Raksha Sharma} \Email{raksha.sharma@cs.iitr.ac.in}\\
\addr Department of Computer Science, IIT Roorkee}

\begin{document}


\maketitle

\begin{abstract}
Large language models (LLMs) are increasingly asked to judge whether a news
article is true, and their stated confidence is often taken at face value. We
test what happens to that confidence when an article's writing style, but not
its facts, is manipulated, as in recent \emph{style attacks} on fake-news
detectors. Tracing how the prediction and its confidence form layer by layer
inside the network, we study three model families (LLaMA-3-8B, Qwen2.5-7B,
Mistral-7B) on three fake-news benchmarks, and find the same three patterns in
every model--dataset pairing. First, on the original articles the models are
competent and largely honest judges: they settle on an answer, briefly become
overconfident, and then correct their confidence in the final layers. Second,
style-rewritten articles break this self-correction: the models become far less
accurate, under the strongest attack, worse than random guessing, yet remain
just as confident, so they are \emph{confidently wrong} precisely when a warning
is most needed. Third, nearly all of the damage comes from rewriting into
sensational or emotionally charged styles, while sober, neutral rewrites leave
the models' honesty largely intact. The practical implication is that an LLM's
confidence about veracity cannot be trusted as a safety signal when an article's
style may have been manipulated.
\end{abstract}


\section{Introduction}
\label{sec:intro}

Automatic fake-news detectors learn to separate real from fake partly through
\emph{style}: fabricated articles often differ stylistically from genuine
reporting. \citet{wu2024sheepdog} show that this is also a vulnerability:
rewriting a fake article into the neutral, authoritative register of a credible
outlet---preserving its false claims---degrades detector F1 by up to roughly
$38\%$. Such rewrites cost cents per article, making the attack cheap and
realistic.

Prior analyses characterize \emph{what} breaks (classifier accuracy) or
\emph{how} representations move geometrically under fine-tuning, but they do not
address the model's \emph{confidence}. This matters for deployment: a detector
that is wrong but uncertain invites human review, whereas one that is wrong and
\emph{certain} does not. The dangerous regime is therefore overconfidence on
attacked inputs, not merely lost accuracy.

We study this regime directly. Taking three off-the-shelf, instruction-tuned LLMs
and prompting them \emph{zero-shot} to classify articles as real or fake, we use
the logit lens \citep{nostalgebraist2020logitlens} to read both the prediction
and the confidence after \emph{every} transformer layer, on original articles and
on style-rewritten ones. The zero-shot setting is deliberate: with no
task-specific training, any calibration behavior reflects the model's intrinsic
veracity judgment rather than an artifact of fine-tuning.

\paragraph{Contributions.}
(i)~We bring layer-wise calibration analysis into the
fake-news adversarial-robustness setting, zero-shot. (ii)~We show a original-data
\emph{confidence-correction phase} exists for this task and replicates across three
model families and three datasets. (iii)~We show style attacks induce
\emph{upper-layer calibration fragility}: the model becomes confidently wrong, and
under the strongest attack accuracy drops \emph{significantly below chance}
(systematic inversion), with the failure co-located with the layers that normally
perform the correction. (iv)~We reproduce, in calibration space and zero-shot, the
reliable-vs-unreliable style asymmetry previously reported for fine-tuned BERT
embeddings, with non-overlapping effect sizes across all nine cells and under two
confidence definitions.

\section{Related Work}
\label{sec:related}

\paragraph{Calibration of LLMs.} Modern neural networks are frequently
miscalibrated \citep{guo2017calibration}, yet large LLMs often exhibit
surprisingly faithful token-level confidence \citep{kadavath2022know}. Confidence
can also be elicited by direct verbalization \citep{tian2023just,xiong2024can},
and calibration is known to degrade under distribution shift
\citep{ovadia2019can}; our attacks are a targeted, label-preserving shift. Recent
mechanistic work localizes confidence regulation to final-layer structures such
as entropy neurons that write into the null space of the unembedding
\citep{stolfo2024confidence}. \citet{joshi2025calibration} show, using the logit
lens, that calibration \emph{evolves across depth}, with a late-layer correction
phase on multiple-choice benchmarks. We extend this layer-wise view from standard
multiple-choice questions to a fake-news setting under adversarial style attacks.

\paragraph{Interpretability via the logit lens and residual stream.} Decoder-only
transformers maintain a residual stream updated additively by each block
\citep{elhage2021mathematical}. The logit lens
\citep{nostalgebraist2020logitlens} projects intermediate residual states through
the unembedding to inspect the model's evolving next-token distribution without
training auxiliary probes, which is the instrument we use throughout. 
\paragraph{Style attacks on fake-news detectors.} \citet{wu2024sheepdog}
introduce SheepDog and a style-attack benchmark in which LLM-generated rewrites
preserve veracity but alter tone; they propose a training-side defense. Their
analysis targets fine-tuned classifiers and their accuracy. We provide a
layer-wise account in \emph{calibration} space and in the \emph{zero-shot}
regime, where no fine-tuning is involved.

\section{Method}
\label{sec:method}

\subsection{Task formulation}
We cast binary veracity detection as a two-option multiple-choice question so
that the answer is a single token amenable to the logit lens:
\begin{quote}\itshape
Article: \texttt{<text>} $\;\ldots\;$ Is the article above real or fake news?
A.~\texttt{<option>} \; B.~\texttt{<option>} \; Answer:
\end{quote}
Placing the article before the question puts the final prompt token at the
answer position. We randomize the mapping of Real/Fake to A/B per article (with a
fixed per-article rule, so an article and all its rewrites share the same
layout), which prevents the well-known positional ``A'' prior from dominating the
calibration measurement. The model is never fine-tuned; all evaluations are
zero-shot.

\subsection{Calibration metrics}
For predictions partitioned into $M$ equal-width confidence bins
$\{B_m\}_{m=1}^{M}$, with per-bin mean confidence $\mathrm{conf}(B_m)$ and
accuracy $\mathrm{acc}(B_m)$, the Expected and Maximum Calibration Errors are
\begin{equation}
\mathrm{ECE}=\sum_{m=1}^{M}\frac{|B_m|}{n}\bigl|\mathrm{acc}(B_m)-\mathrm{conf}(B_m)\bigr|,
\qquad
\mathrm{MCE}=\max_{m}\bigl|\mathrm{acc}(B_m)-\mathrm{conf}(B_m)\bigr|.
\end{equation}
Both vanish for a perfectly calibrated model; we use $M=10$.

\subsection{Layer-wise read-out via the logit lens}
Let $h_\ell$ denote the residual stream at layer $\ell$ for the final prompt
token, $\mathrm{LN}(\cdot)$ the model's final normalization, and $W_U$ the
unembedding. The logit lens read-out and its distribution are
\begin{equation}
z_\ell = W_U\,\mathrm{LN}(h_\ell),
\qquad
p_\ell = \mathrm{softmax}(z_\ell).
\end{equation}
A single forward pass yields $h_\ell$ for all $\ell$; the top read-out reproduces
the model's true output logits. We report two confidence definitions:
\emph{paper-faithful} (argmax over the full vocabulary and its probability,
matching \citealp{joshi2025calibration}) and \emph{restricted} (probability over
$\{$A,\,B$\}$, renormalized). The two measures agree closely on all large
effects---final-layer original-test ECE, test-level $\Delta$ECE, and unreliable-style
$\Delta$ECE differ by at most ${\sim}0.05$ across cells, and the
reliable/unreliable non-overlap holds under both (Appendix~\ref{apd:variants})---and differ only on
the small reliable-style effect, whose sign is not robust; we therefore never
rest a claim on it. We present paper-faithful and note that, in the \emph{middle}
layers, its top token is often an ordinary word rather than an answer letter, so
the informative signal lies in the upper, answer-predicting layers.

\subsection{Robustness quantities}
At the final layer we report, relative to the original condition: the
\emph{Accuracy Collapse Ratio} $\mathrm{AccCR}=\mathrm{Acc}_{\text{att}}/\mathrm{Acc}_{\text{orig}}$
($1.0$ means no damage), and the calibration shift
$\Delta\mathrm{ECE}=\mathrm{ECE}_{\text{att}}-\mathrm{ECE}_{\text{orig}}$
(large positive values indicate the model became overconfident). The
\emph{correction magnitude} is the upper-layer rise-then-fall in ECE on original
data. We average these separately over the two reliable styles (objective,
neutral) and the two unreliable styles (sensational, emotionally triggering).

\subsection{Statistical reliability}
We report Wilson $95\%$ confidence intervals for accuracy and, via the Katz log
method, for the accuracy collapse ratio; these are exact functions of the reported
proportions and sample sizes. The four adversarial sets restyle the \emph{same}
test articles and are therefore not independent, so we treat the effective sample
size as the test-set size and report per-set intervals rather than pooling.
Aggregated ``Adv'' figures average per-set metrics (i.e.\ averaged ECE, not the
ECE of a pooled set).

\subsection{Models and data}
We evaluate three instruction-tuned decoder-only models from different families:
\textbf{LLaMA-3-8B-Instruct} ($32$ layers; \citealp{dubey2024llama3}),
\textbf{Qwen2.5-7B-Instruct} ($28$ layers; \citealp{qwen2024qwen25}), and
\textbf{Mistral-7B-Instruct} ($32$ layers; \citealp{jiang2023mistral}). We use the
three SheepDog benchmarks \citep{wu2024sheepdog}: \textbf{PolitiFact} and
\textbf{GossipCop} from FakeNewsNet \citep{shu2020fakenewsnet}, and \textbf{LUN}
\citep{rashkin2017truth}. Each ships original articles, four GPT-3.5 style
reframings of each training article (objective, neutral, sensational,
emotionally triggering), and four adversarial test sets (adv A to D).\\ 
\citep{wu2024sheepdog} establishes a set of four general style-oriented adjectives for the prompt: “objective and professional” and “neutral” to emulate reliable sources, and “emotionally triggering” and “sensational” for unreliable sources.
\\
\textbf{Rewrite the following article in a / an [specified] tone: [news article]} \\

LLM style-attack framework of \citet{wu2024sheepdog}, which rewrites articles using the prompt:

\textbf{Rewrite the following article using the style of [publisher name]:} \textit{[news article]}

Two transformations are applied:
\begin{itemize}
    \item \textbf{Disguising fake news} as reputable publishers (CNN, The New York Times), raising perceived credibility.
    \item \textbf{Disguising real news} as tabloid publishers (National Enquirer, The Sun), lowering perceived credibility.
\end{itemize}
This yields four adversarial test sets labeled A to D (Table~\ref{tab:adversarial_sets}).

\begin{table}[h]
\centering
\begin{tabular}{ccc}
\toprule
Publisher name & CNN & The New York  \\
            &  &  Times\\
\midrule
National Enquirer & A & B \\
The Sun & C & D \\
\bottomrule
\end{tabular}
\caption{Configuration of the four style-based adversarial test sets from \citet{wu2024sheepdog}.}
\label{tab:adversarial_sets}
\end{table}
 LUN ($1{,}500$ test / $6{,}000$ train) and GossipCop ($1{,}584$ / $6{,}332$) are
large; PolitiFact ($90$ / $360$) is small in size.
\section{Results}
\label{sec:results}

\begin{figure}[t]
\centering
\includegraphics[width=\textwidth]{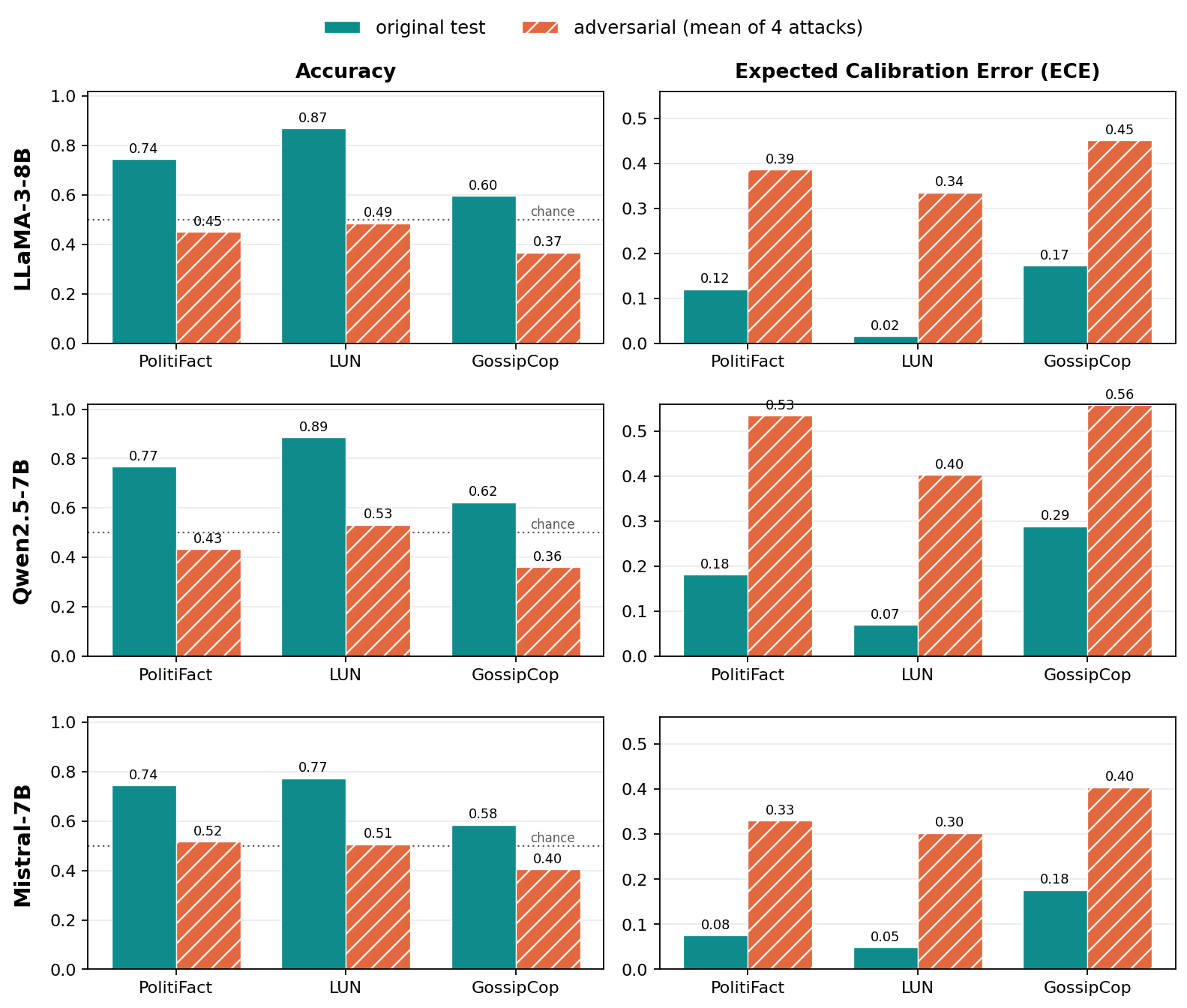}
\caption{\textbf{Original test versus adversarial at the final layer, for every
model (rows) and dataset.} Left column: accuracy (dotted line marks the $0.5$
chance level); right column: ECE (lower is better). Teal = original test; hatched
coral =
adversarial (averaged over the four attack sets). In every case the attack drives
accuracy toward or below chance while ECE rises; it's the signature of a model that
stays confident as it becomes wrong. (Bars average the four attacks; the
\emph{strongest} attack falls further---below chance in every case,
Table~\ref{tab:acc_ci}.) Mistral-7B is the most robust (its
adversarial accuracy bars are highest and its adversarial ECE lowest) yet exhibits
the same pattern.}
\label{fig:dashboard}
\end{figure}

All three findings hold true across all nine model-dataset combinations; specifically, the original-data confidence-correction phase is consistently detected in all cells.
Figure~\ref{fig:dashboard} is the quickest way to read the entire study: it shows,
for every model and data set at the final layer, the original-test result alongside the
adversarial one. Across all nine cells, the accuracy collapses towards or below chance
, while ECE rises sharply, and the model remains confident as it becomes wrong. The
per-layer figures that follow (Figures~\ref{fig:adv}--\ref{fig:reframe}) show how
these end states arise across depth, and Table~\ref{tab:master} lists the numbers.

\subsection{A confidence-correction phase on original inputs}
On original articles, accuracy is near zero until the answer emerges in the upper
layers (around layer $22$ of $32$ for the $32$-layer models, proportionally
similar for Qwen's $28$), then saturates; confidence transiently overshoots
accuracy (raising ECE); and the final layers recalibrate confidence downward. The
completeness of this correction tracks task difficulty: on LUN, LLaMA ends at
$0.869$ accuracy with ECE $0.016$ (near-perfect), whereas on the harder GossipCop
set it ends less calibrated (ECE $0.173$ for LLaMA, $0.289$ for Qwen, $0.176$ for
Mistral). 

\subsection{Style attacks induce confident error in the upper layers}
\begin{figure}[t]
\centering
\includegraphics[width=\textwidth]{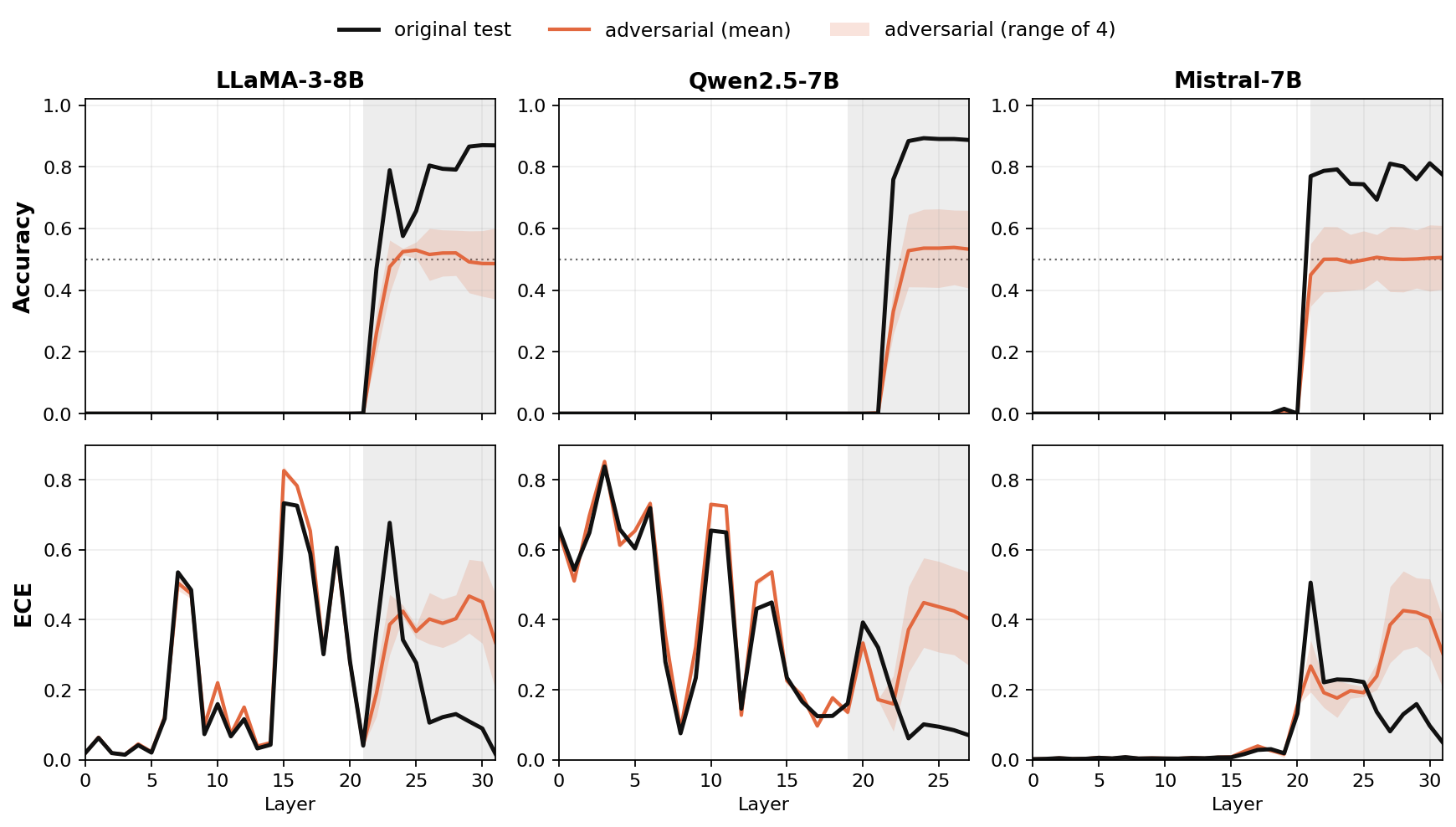}
\caption{\textbf{Original test versus adversarial across depth, on LUN, for all
three models} (columns). Top row: accuracy per layer (dotted line marks the $0.5$
chance level); bottom row: ECE per layer. Black = original test; coral line = mean
over the four adversarial sets, with the shaded band spanning their range; the
grey vertical band marks the answer-predicting upper layers. In every model the
original-test curve saturates well above chance and recalibrates ECE downward,
whereas under attack accuracy is pinned at or below chance while ECE stays
high---the signature of confident error. Mistral-7B (right) is the most robust but
shows the same shape.}
\label{fig:adv}
\end{figure}

Averaged over the four adversarial sets, final-layer accuracy collapses to
$0.56$--$0.69$ of the original value (Table~\ref{tab:master}, AccCR), while confidence remains
high, so $\Delta\mathrm{ECE}$ is $+0.23$ to $+0.35$ in every case. Because LUN's
original-test baseline is so well calibrated ($0.016$), the attack converts a
near-perfect detector into a badly overconfident one, and the same pattern recurs
on Qwen and Mistral. Mistral-7B is the most robust of the three (the mildest
collapse, AccCR up to $0.69$ and $\Delta\mathrm{ECE}$ as low as $+0.23$), yet
exhibits every signature---a robustness \emph{gradient} across families with an
invariant qualitative pattern. The accuracy collapse is statistically unambiguous
(AccCR $95\%$ upper bounds lie well below $1$ in all cases,
Table~\ref{tab:acc_ci}), though on PolitiFact ($n{=}90$) the intervals are wide
and we treat its per-cell magnitudes as indicative. Under the attacks, accuracy
climbs where the answer first forms but stalls at or below chance through the upper
layers while confidence stays near $0.95$, and this shape is the same in all three
models (Figure~\ref{fig:adv}).

\paragraph{Below-chance behavior.} The collapse is more than lost signal. The
most damaging of the four adversarial sets drives final-layer accuracy
\emph{significantly below} the $0.5$ chance level in \emph{every} one of the nine
cases (Wilson $95\%$ upper bound $<0.5$; Table~\ref{tab:acc_ci}), including
PolitiFact at $n{=}90$ ($0.33$, CI $[0.25,0.44]$); on GossipCop even the four-set The 
average is sub-chance. Below-chance accuracy on a balanced binary task means the
detector is worse than random---confidently emitting the wrong label---not merely
uncertain: the attack pushes the model past a coin flip, not toward it. Because the
attack restyles both classes, this is consistent with \emph{systematic inversion};
distinguishing genuine inversion from strongly one-sided error (e.g.\ camouflaged
fakes alone) would require the per-class confusion matrix, which we leave to future
work.

\begin{table}[t]
\centering
\footnotesize
\setlength{\tabcolsep}{4pt}
\caption{Accuracy with Wilson $95\%$ confidence intervals (final layer,
paper-faithful). ``Strongest adv'' is the most damaging of the four adversarial
sets; its interval upper bound is below the $0.5$ chance level in every row, so
under the strongest attack the model is significantly \emph{worse than chance}
(consistent with systematic inversion; see text). AccCR intervals (Katz log) lie
below $1$ throughout,
confirming the collapse; PolitiFact's width reflects $n{=}90$.}
\label{tab:acc_ci}
\begin{tabular}{l c c c}
\toprule
Model / Dataset & Original test acc & Strongest adv acc & AccCR \\
\midrule
LLaMA-3-8B / PolitiFact & 0.744 [0.65,\,0.82] & 0.333 [0.25,\,0.44] & 0.604 [0.47,\,0.78] \\
LLaMA-3-8B / LUN        & 0.869 [0.85,\,0.89] & 0.371 [0.35,\,0.40] & 0.559 [0.53,\,0.59] \\
LLaMA-3-8B / GossipCop  & 0.596 [0.57,\,0.62] & 0.328 [0.31,\,0.35] & 0.616 [0.57,\,0.67] \\
\midrule
Qwen2.5-7B / PolitiFact & 0.767 [0.67,\,0.84] & 0.322 [0.24,\,0.42] & 0.565 [0.44,\,0.74] \\
Qwen2.5-7B / LUN        & 0.887 [0.87,\,0.90] & 0.407 [0.38,\,0.43] & 0.601 [0.57,\,0.63] \\
Qwen2.5-7B / GossipCop  & 0.623 [0.60,\,0.65] & 0.314 [0.29,\,0.34] & 0.581 [0.54,\,0.63] \\
\midrule
Mistral-7B / PolitiFact & 0.744 [0.65,\,0.82] & 0.367 [0.27,\,0.47] & 0.694 [0.55,\,0.88] \\
Mistral-7B / LUN        & 0.773 [0.75,\,0.79] & 0.403 [0.38,\,0.43] & 0.654 [0.62,\,0.69] \\
Mistral-7B / GossipCop  & 0.585 [0.56,\,0.61] & 0.357 [0.33,\,0.38] & 0.691 [0.64,\,0.74] \\
\bottomrule
\end{tabular}
\end{table}

\subsection{The damage is driven by unreliable styles}
\begin{figure}[t]
\centering
\includegraphics[width=\textwidth]{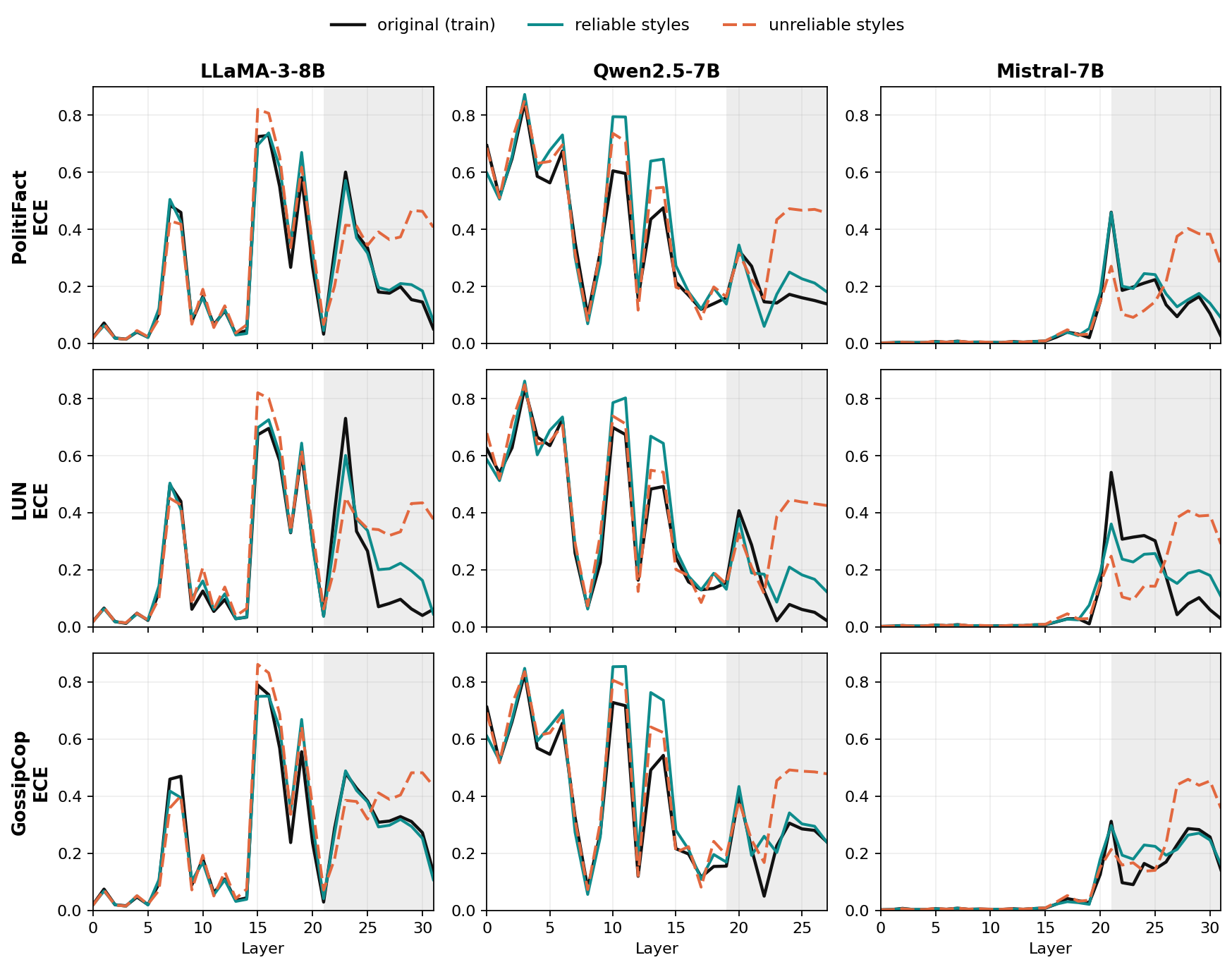}
\caption{\textbf{Reliable versus unreliable reframings: ECE across depth for all
nine model$\times$dataset cells} (rows = datasets, columns = models;
article-matched). Black = original (train); teal = mean of the reliable styles
(objective, neutral); dashed coral = mean of the unreliable styles (sensational,
emotionally triggering). In every cell the reliable curve tracks the original
baseline through the answer-bearing upper layers (grey band), while the unreliable
curve diverges upward---the calibration damage is style-selective in every model
and dataset. Colour and line style distinguish the groups for grayscale.
Corresponding accuracy trajectories are in Appendix~\ref{apd:morefigs}.}
\label{fig:reframe}
\end{figure}

Comparing each article to its four style-matched rewrites isolates the effect of
style. Reliable rewrites (objective, neutral) leave calibration essentially
intact: across all nine cases their $\Delta\mathrm{ECE}\in[-0.026,\,+0.100]$, and
on GossipCop they even slightly raise accuracy ($\mathrm{AccCR}\approx1.02$).
Unreliable rewrites (sensational, emotionally triggering) are damaging:
$\Delta\mathrm{ECE}\in[+0.216,\,+0.404]$, with accuracy falling to
$0.58$--$0.83$ of the original value. The two ranges \emph{do not overlap}---the worst
reliable-style damage ($+0.10$) is milder than the gentlest unreliable-style
damage ($+0.22$)---so the effect is closer to a switch than a gradient.
Figure~\ref{fig:reframe} shows the ECE trajectories for \emph{all nine cases}: in
every dataset and model the reliable curves track the original baseline while the
unreliable curves diverge in the upper layers (the corresponding accuracy
trajectories are in Appendix~\ref{apd:morefigs}). A plausible
mechanism is
\emph{stylistic homogeneity}: GPT-3.5's sensational/emotional rewrites share an
alarmist register that makes all articles read as fake-like, collapsing the cues
the model uses to separate the classes. We caution that this asymmetry is partly a
property of the \emph{rewrites}: all reframings come from a single generator
(GPT-3.5), whose sensational/emotional outputs may be especially homogeneous. The
pattern may thus reflect the attacker's text
distribution as much as the detector's reasoning; the non-overlap nonetheless
holds under both confidence measures (Appendix~\ref{apd:variants}).

\subsection{A preliminary observation on the attack sets}
On the cells we inspected per set, adversarial sets A and B are consistently more
damaging than C and D (on LLaMA/LUN, adv-A/B reach $\approx0.37$ accuracy versus
$\approx0.59$ for adv-C/D, with the same ordering on PolitiFact and GossipCop).

\begin{table}[t]
\centering
\footnotesize
\setlength{\tabcolsep}{3.5pt}
\caption{Final-layer results (paper-faithful confidence; restricted agrees on all
large effects, Appendix~\ref{apd:variants}). ECE: lower is better. AccCR: $1.0$ =
no accuracy damage. $\Delta$ECE: higher = worse calibration after attack. ``Adv''
columns average per-set metrics over the four adversarial sets (not pooled);
``rel.''/``unrel.'' average the two reliable/unreliable reframings. The original-data
confidence-correction phase is present in all nine rows.}
\label{tab:master}
\begin{tabular}{l c c c c c c}
\toprule
Model / Dataset & Orig.\ Acc & Orig.\ ECE & Adv AccCR & Adv $\Delta$ECE & rel.\ $\Delta$ECE & unrel.\ $\Delta$ECE \\
\midrule
LLaMA-3-8B / PolitiFact & 0.744 & 0.120 & 0.604 & $+0.266$ & $+0.027$ & $+0.355$ \\
LLaMA-3-8B / LUN        & 0.869 & \textbf{0.016} & 0.559 & $+0.319$ & $-0.016$ & $+0.316$ \\
LLaMA-3-8B / GossipCop  & 0.596 & 0.173 & 0.616 & $+0.278$ & $-0.026$ & $+0.302$ \\
\midrule
Qwen2.5-7B / PolitiFact & 0.767 & 0.182 & 0.565 & $+0.353$ & $+0.042$ & $+0.319$ \\
Qwen2.5-7B / LUN        & 0.887 & 0.070 & 0.601 & $+0.333$ & $+0.100$ & $+0.404$ \\
Qwen2.5-7B / GossipCop  & 0.623 & 0.289 & 0.581 & $+0.269$ & $-0.002$ & $+0.239$ \\
\midrule
Mistral-7B / PolitiFact & 0.744 & 0.075 & 0.694 & $+0.255$ & $+0.064$ & $+0.247$ \\
Mistral-7B / LUN        & 0.773 & 0.048 & 0.654 & $+0.254$ & $+0.078$ & $+0.262$ \\
Mistral-7B / GossipCop  & 0.585 & 0.176 & 0.691 & $+0.228$ & $+0.018$ & $+0.216$ \\
\bottomrule
\end{tabular}
\end{table}

\section{Discussion}
\label{sec:discussion}

The results cohere into one account. On original texts these LLMs are competent
veracity judges with faithful confidence, achieved through a late-layer
correction. That judgment, however, appears to depend on stylistic cues, so an
emotionally charged rewrite is associated with a confident, incorrect verdict
forming in the upper layers. We emphasize that the zero-shot setting matters
here: an encoder with no task training never forms a veracity judgment at all, whereas our zero-shot decoders do, and that judgment is
fragile. One important alternative we cannot yet exclude is that unreliable rewrites are simply more out-of-distribution and would degrade the model on
\emph{any} task, not veracity specifically; separating general OOD degradation
from veracity-specific fragility requires a same-articles control task (e.g.\
topic classification on the identical rewrites) and is the single most important
experiment we leave to future work. Subject to that caveat, the fragility seems
intrinsic to how a capable model judges whether text ``sounds true,'' rather than
a byproduct of task-specific training. Practically, an LLM's confidence on
potentially restyled content should not be used as a decision gate; emotionally
styled inputs warrant additional scrutiny.

\section{Conclusion}
\label{sec:conclusion}

Reading per-layer prediction and confidence from three zero-shot LLMs across three
fake-news benchmarks, we find an original-data confidence-correction phase that style
rewriting suppresses: accuracy collapses while confidence stays high,under the
strongest attack, past chance into worse-than-random behavior, and the damage is
associated almost entirely with sensational and emotional rewriting. All patterns
hold in all nine model$\times$dataset cells, with reliable and unreliable calibration
effects that do not overlap under either confidence measure.
The
practical implication is clear: an LLM's confidence about veracity is unreliable
when an article's style may have been manipulated.

\acks{Acknowledgements (omitted for review).}

\bibliography{refs}

\appendix
\section{Reproducibility details}
\label{apd:repro}
All models are evaluated in \texttt{bfloat16}. Confidence is read at the final
prompt token; intermediate read-outs apply the model's final normalization to the
raw block output, except the top read-out, which uses the already-normalized
final hidden state (so it reproduces the model's true output logits). We use
$M{=}10$ calibration bins, a fixed random seed for the per-article Real/Fake
option assignment, and truncate article bodies (preserving the lead) so the
question and answer slot are never removed. For the article-matched contrast,
reframing files are row-aligned to the original training articles and inherit
their labels; for the test-level contrast, adversarial sets are row-aligned to
the original test set.


\paragraph{Confidence intervals.} Accuracy intervals are Wilson score intervals;
AccCR intervals use the Katz log method. Both use the test-set size as the
effective $n$; because the four adversarial sets restyle the same articles, we do
not pool them.

\section{Paper-faithful vs.\ restricted confidence}
\label{apd:variants}
Table~\ref{tab:variants} compares the two confidence definitions at the final
layer. They agree on all large effects: original-test ECE, test-level $\Delta$ECE, and
unreliable-style $\Delta$ECE differ by at most ${\sim}0.05$, and the
reliable/unreliable non-overlap holds under both (restricted: reliable
$\Delta\mathrm{ECE}\in[-0.03,+0.12]$ vs.\ unreliable $\in[+0.21,+0.38]$). They
diverge only on the small reliable-style effect, whose sign is not robust (e.g.\
$-0.016$ vs.\ $+0.007$ on LLaMA/LUN); we never rest a claim on its sign.

\begin{table}[h]
\centering
\footnotesize
\setlength{\tabcolsep}{4pt}
\caption{Final-layer paper-faithful (pf) vs.\ restricted (re) confidence;
values are pf~/~re. Large effects agree to within ${\sim}0.05$.}
\label{tab:variants}
\begin{tabular}{l c c c c}
\toprule
Model / Dataset & Orig.\ ECE & Test $\Delta$ECE & rel.\ $\Delta$ECE & unrel.\ $\Delta$ECE \\
\midrule
LLaMA-3-8B / PolitiFact & 0.120 / 0.123 & $+0.266$ / $+0.272$ & $+0.027$ / $+0.016$ & $+0.355$ / $+0.352$ \\
LLaMA-3-8B / LUN        & 0.016 / 0.021 & $+0.319$ / $+0.332$ & $-0.016$ / $+0.007$ & $+0.316$ / $+0.345$ \\
LLaMA-3-8B / GossipCop  & 0.173 / 0.187 & $+0.278$ / $+0.279$ & $-0.026$ / $-0.027$ & $+0.302$ / $+0.299$ \\
\midrule
Qwen2.5-7B / PolitiFact & 0.182 / 0.209 & $+0.353$ / $+0.339$ & $+0.042$ / $+0.057$ & $+0.319$ / $+0.300$ \\
Qwen2.5-7B / LUN        & 0.070 / 0.096 & $+0.333$ / $+0.335$ & $+0.100$ / $+0.117$ & $+0.404$ / $+0.377$ \\
Qwen2.5-7B / GossipCop  & 0.289 / 0.328 & $+0.269$ / $+0.264$ & $-0.002$ / $+0.022$ & $+0.239$ / $+0.213$ \\
\midrule
Mistral-7B / PolitiFact & 0.075 / 0.113 & $+0.255$ / $+0.271$ & $+0.064$ / $+0.030$ & $+0.247$ / $+0.256$ \\
Mistral-7B / LUN        & 0.048 / 0.101 & $+0.254$ / $+0.253$ & $+0.078$ / $+0.102$ & $+0.262$ / $+0.305$ \\
Mistral-7B / GossipCop  & 0.176 / 0.223 & $+0.228$ / $+0.230$ & $+0.018$ / $+0.006$ & $+0.216$ / $+0.217$ \\
\bottomrule
\end{tabular}
\end{table}

\section{Per-layer figures for the remaining cells}
\label{apd:morefigs}
\begin{figure}[tbp]
\centering
\includegraphics[width=0.85\textwidth]{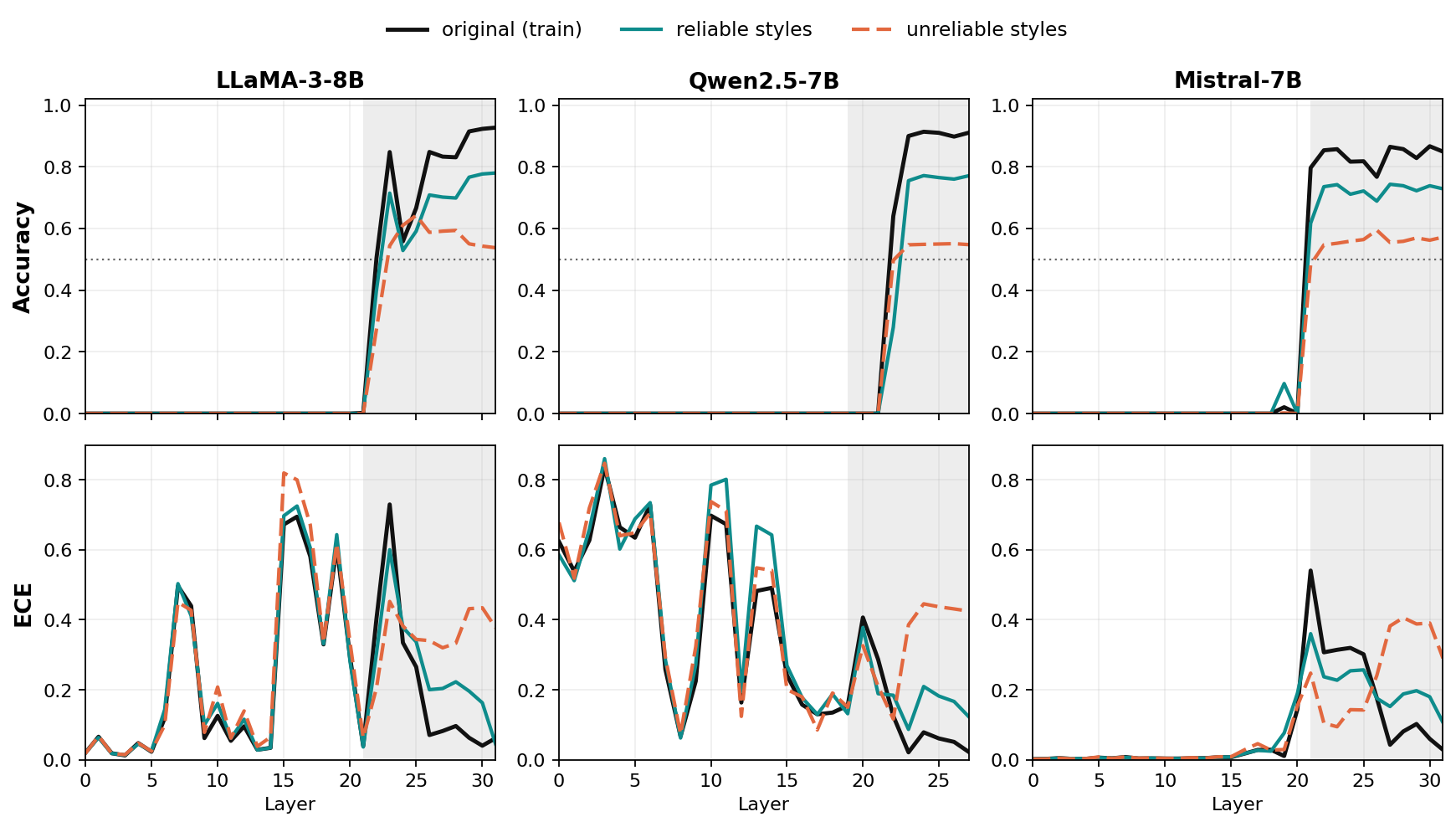}
\caption{\textbf{LUN}: reliable versus unreliable reframings across depth, all
three models---accuracy (top) and ECE (bottom). Black = original (train); teal =
reliable styles; dashed coral = unreliable styles. Accuracy companion to the LUN
row of Figure~\ref{fig:reframe}.}
\label{fig:reframe_lun}
\end{figure}

\begin{figure}[tbp]
\centering
\includegraphics[width=0.85\textwidth]{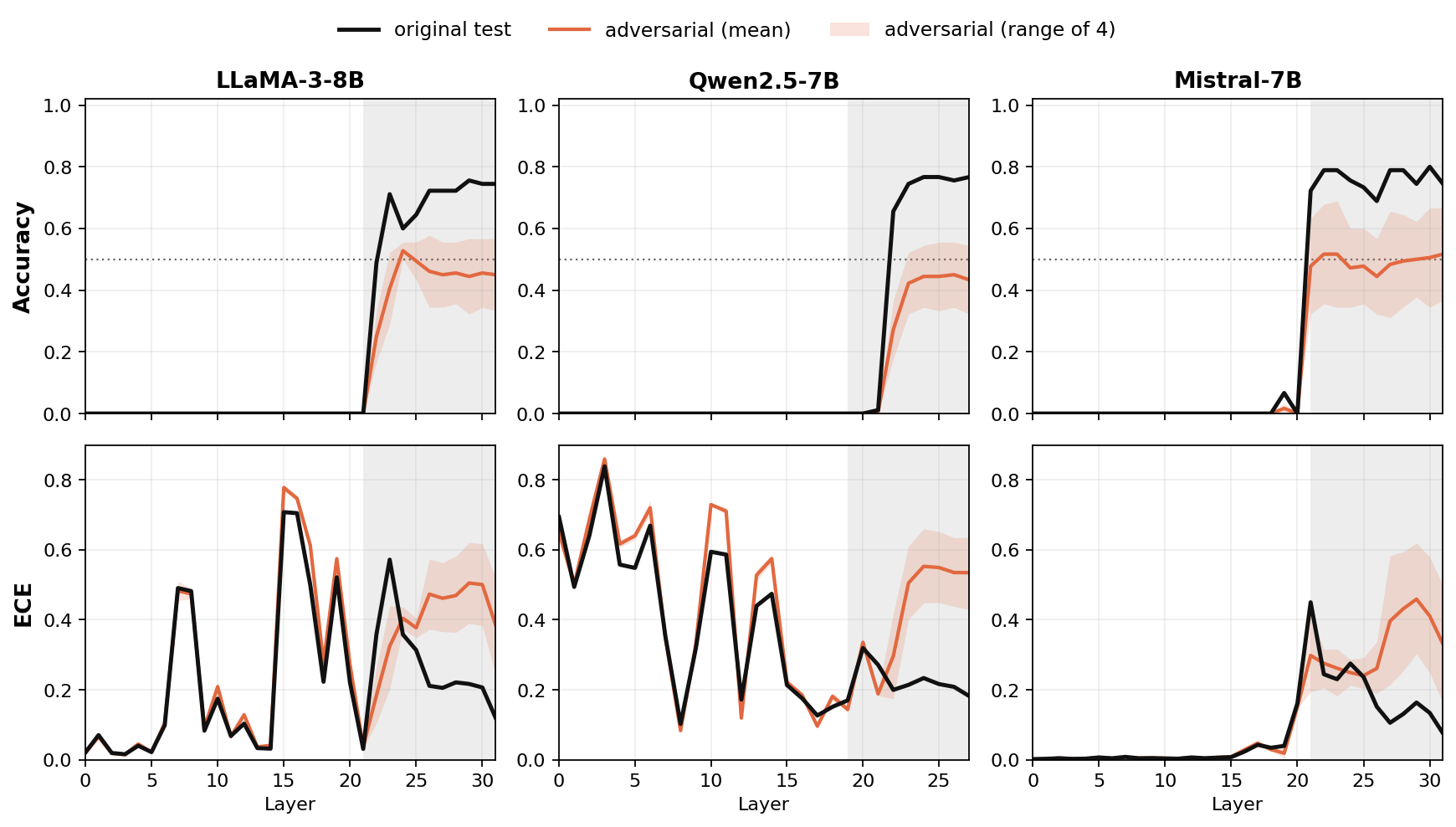}
\caption{\textbf{PolitiFact} ($n{=}90$): original test versus adversarial across
depth, all three models (cf.\ Figure~\ref{fig:adv}). Black = original test; coral
= mean of the four attacks; shaded band = their range; grey band = upper layers.}
\label{fig:adv_pf}
\end{figure}

\begin{figure}[tbp]
\centering
\includegraphics[width=0.85\textwidth]{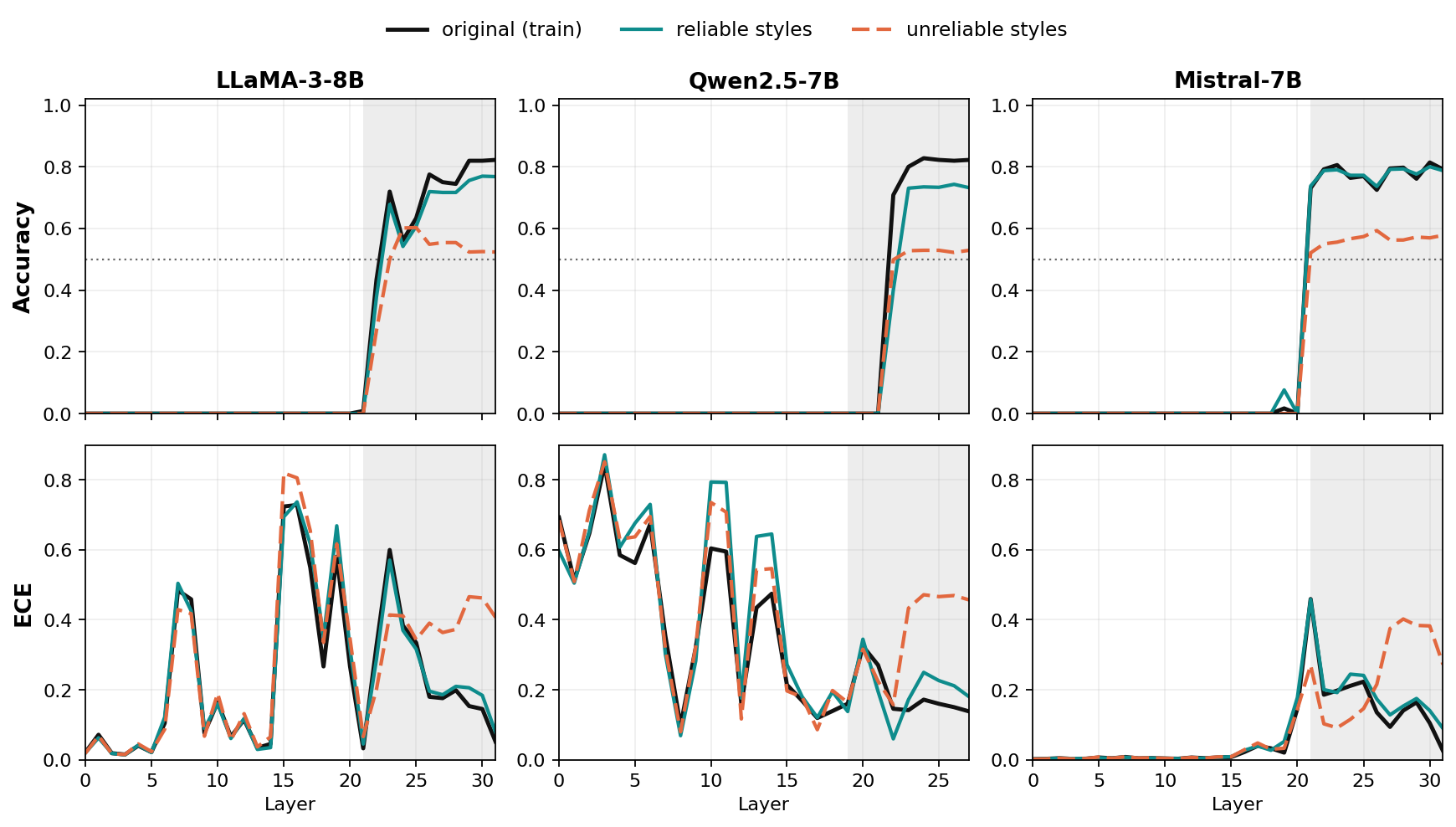}
\caption{\textbf{PolitiFact}: reliable versus unreliable reframings across depth,
all three models (cf.\ Figure~\ref{fig:reframe}). Black = original (train); teal =
reliable styles; dashed coral = unreliable styles.}
\label{fig:reframe_pf}
\end{figure}

\begin{figure}[tbp]
\centering
\includegraphics[width=0.85\textwidth]{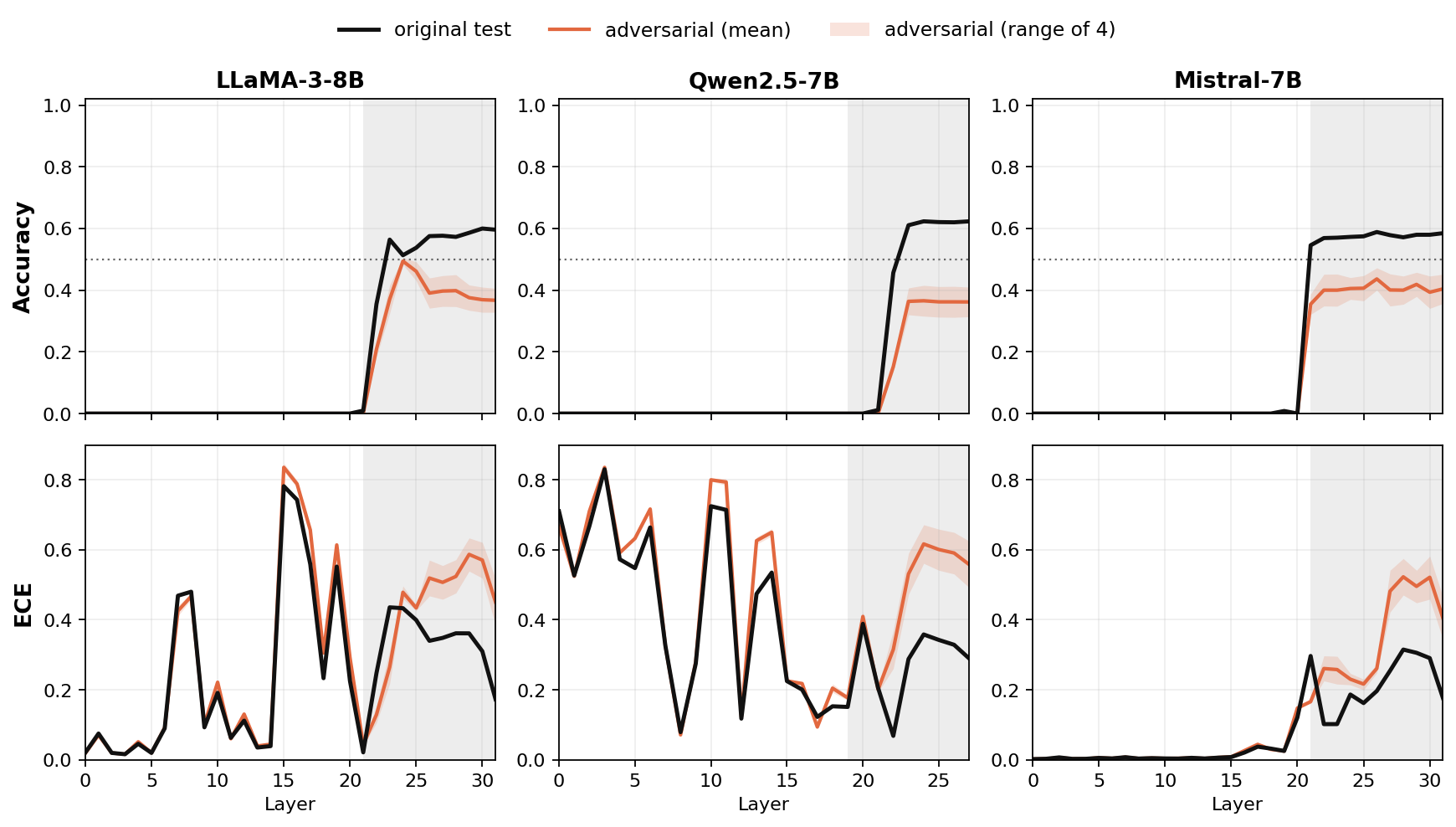}
\caption{\textbf{GossipCop}: original test versus adversarial across depth, all
three models (cf.\ Figure~\ref{fig:adv}). Even though the original-test baseline
is less calibrated on this harder dataset, the attack drives accuracy well below
chance in the upper layers while ECE rises.}
\label{fig:adv_gc}
\end{figure}

\begin{figure}[tbp]
\centering
\includegraphics[width=0.85\textwidth]{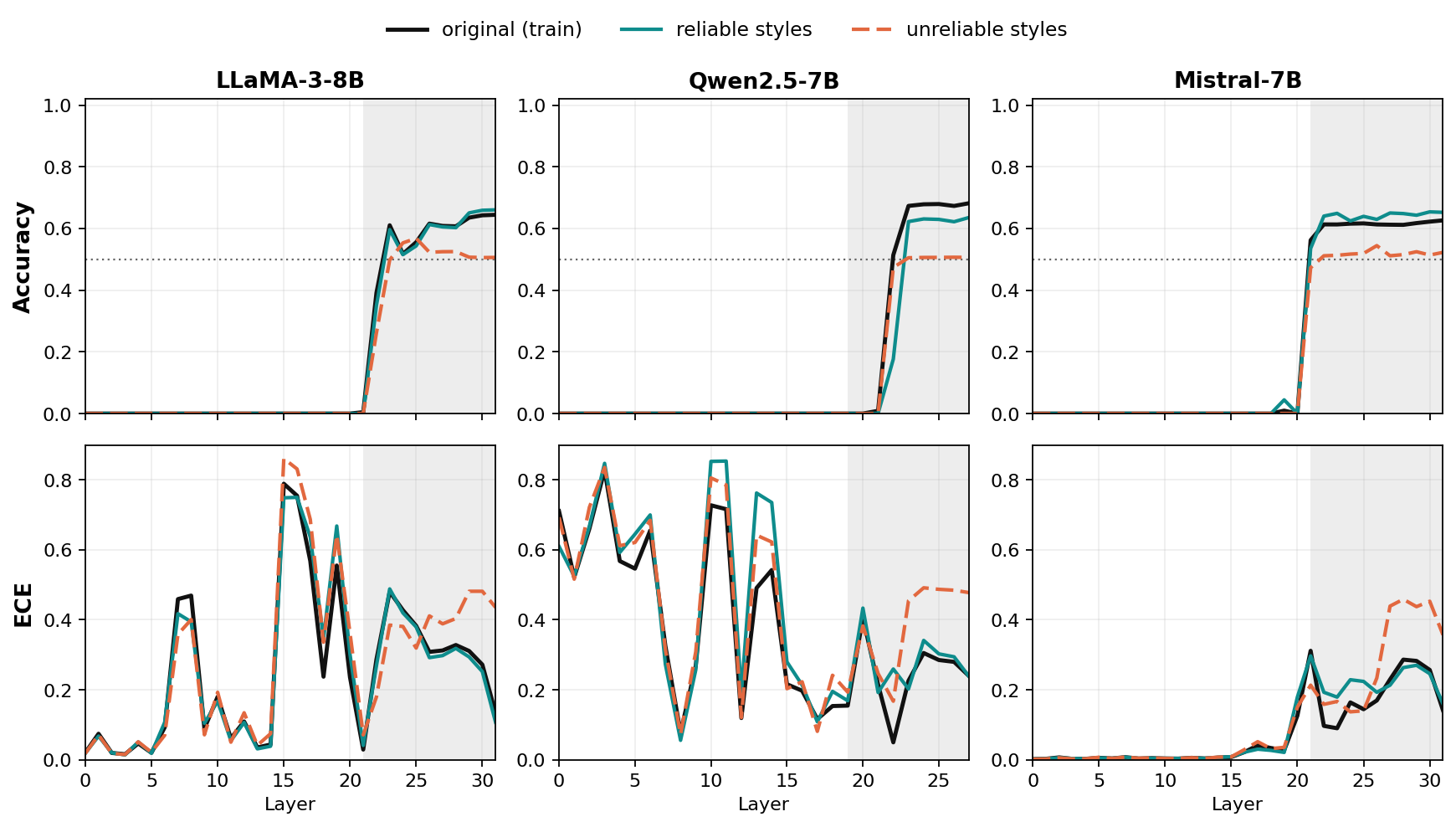}
\caption{\textbf{GossipCop}: reliable versus unreliable reframings across depth,
all three models (cf.\ Figure~\ref{fig:reframe}). Reliable styles track the original baseline;
unreliable styles diverge in the upper layers.}
\label{fig:reframe_gc}
\end{figure}

Figure~\ref{fig:adv} in the main text shows the adversarial contrast across depth
on LUN, and Figure~\ref{fig:reframe} summarizes the reframing contrast (ECE) for
all nine cells. This appendix completes the per-layer picture with the full
accuracy-and-ECE views: the reframing contrast on LUN
(Figure~\ref{fig:reframe_lun}), and both contrasts on PolitiFact
(Figures~\ref{fig:adv_pf} and~\ref{fig:reframe_pf}) and GossipCop
(Figures~\ref{fig:adv_gc} and~\ref{fig:reframe_gc}). The same qualitative pattern
holds throughout: under attack, accuracy is pinned at or below chance through the
upper layers while ECE stays high, and reliable reframings track the original
baseline while unreliable ones diverge. PolitiFact's adversarial bands are wider
because its test set is small ($n{=}90$); GossipCop is the hardest task, so its
original-test baseline is itself less calibrated, yet the attack still drives
accuracy well below chance.


\clearpage 


\clearpage 


\end{document}